\documentclass[preprint,11pt]{elsarticle}

\usepackage{caption}
\usepackage{setspace}
\usepackage{amssymb}
\usepackage{multirow}
\usepackage{booktabs}
\usepackage{amsthm}
\usepackage{booktabs}
\usepackage{amsmath}
\usepackage{amssymb}
\usepackage{mathrsfs}
\usepackage{longtable}
\usepackage{lscape}
\usepackage{url}
\usepackage{tabularx}
\usepackage{epsfig}
\usepackage{epstopdf}
\usepackage{colortbl}
\usepackage{subfigure}
\usepackage{graphicx}
\usepackage{footnote}
\usepackage{tikz,mathpazo}
\usepackage{makecell}
\newtheorem{example}{Example}[section]

\newtheorem{definition}{Definition}[section]
\newtheorem{property}{Property}[section]
\newtheorem{remark}{Remark}[section]

\newcommand{\ie}{\textit{i}.\textit{e}.}

\usetikzlibrary{shapes.geometric, arrows}
\usepackage{geometry}
\begin{document}
\begin{frontmatter}

\title{Repeatable Random Permutation Set}



\author[address1]{Wenran Yang}
\author[address1,vt,jt,eth]{Yong Deng\corref{label1}}

\cortext[label1]{Corresponding author: Yong Deng, Institute of Fundamental and Frontier Science, University of Electronic Science and Technology of China, Chengdu, 610054, China
Email address: dengentropy@uestc.edu.cn; prof.deng@hotmail.com }

\address[address1]{Institute of Fundamental and Frontier Science, University of Electronic Science and Technology of China, Chengdu, 610054, China}
\address[vt]{School of Education, Shannxi Normal University, Xi'an, China}
\address[jt]{School of Knowledge Science, Japan Advanced Institute of Science and Technology, Nomi, Ishikawa 923-1211, Japan}
\address[eth]{Department of Management, Technology, and Economics, ETH Zurich, Zurich, Switzerland}
\begin{abstract} 
Random permutation set (RPS), as a recently proposed theory, enables powerful information representation by traversing all possible permutations. 
However, the repetition of items is not allowed in RPS while it is quite common in real life. To address this issue, we propose repeatable random permutation set ($\rm R^2PS$) which takes the repetition of items into consideration. The right and left junctional sum combination rules are proposed and their properties including consistency, pseudo-Matthew effect and associativity are researched. Based on these properties, a decision support system application is simulated to show the effectiveness of $\rm R^2PS$. 

\end{abstract}


\end{frontmatter}

\section{Introduction}
\label{Introduction}
Since Uncertainty is ubiquitous in real life, many theories are proposed to tackle uncertainty. Among them, Dempster-Shafer evidence theory (DST) \cite{dempster2008upper,shafer2021mathematical} has received wide attention and successfully applied in many fields, such as  knowledge representation \cite{limboo2022q} and information fusion \cite{Xiao2022Generalizeddivergence}.
From the time it was proposed, to the present DST has been deeply researched, including uncertainty measurement \cite{deng2016deng,deng2020uncertainty,BALAKRISHNAN2022unified} ,complex evidence theory \cite{Xiao2021CEQD,Xiao2022NQMF} and distribution features \cite{deng2022Powerlaw}. 


Random permutation set (RPS) is a recently proposed uncertainty processing theory \cite{deng2022random}. In DST, given a frame of discernment with $N$ hypothesis, all combinations of hypothesizes consist a $2^N$ power set, which is the research object of DST. RPS researches all permutations of hypothesizes instead of combinations. In RPS, all possible permutations are traversed and assigned belef. Due to the enlargement of event space, the ability for information represetation is powerful. 
The information volume \cite{deng2022RPSentropy, deng2021maximum}, diversity measures \cite{deng2022RPSdistance} and combination rules \cite{deng2022random} of RPS are also researched. 

In RPS, the permutations of hypothesizes are sequences without repetition. However, in real life, the recurrence of elements is quite common, such as currency denominations and nitrogenous bases of DNA. Consequently, it is necessary to propose a model of permutation with repetition. 

\begin{figure}
	\centering
	\includegraphics[height=8cm]{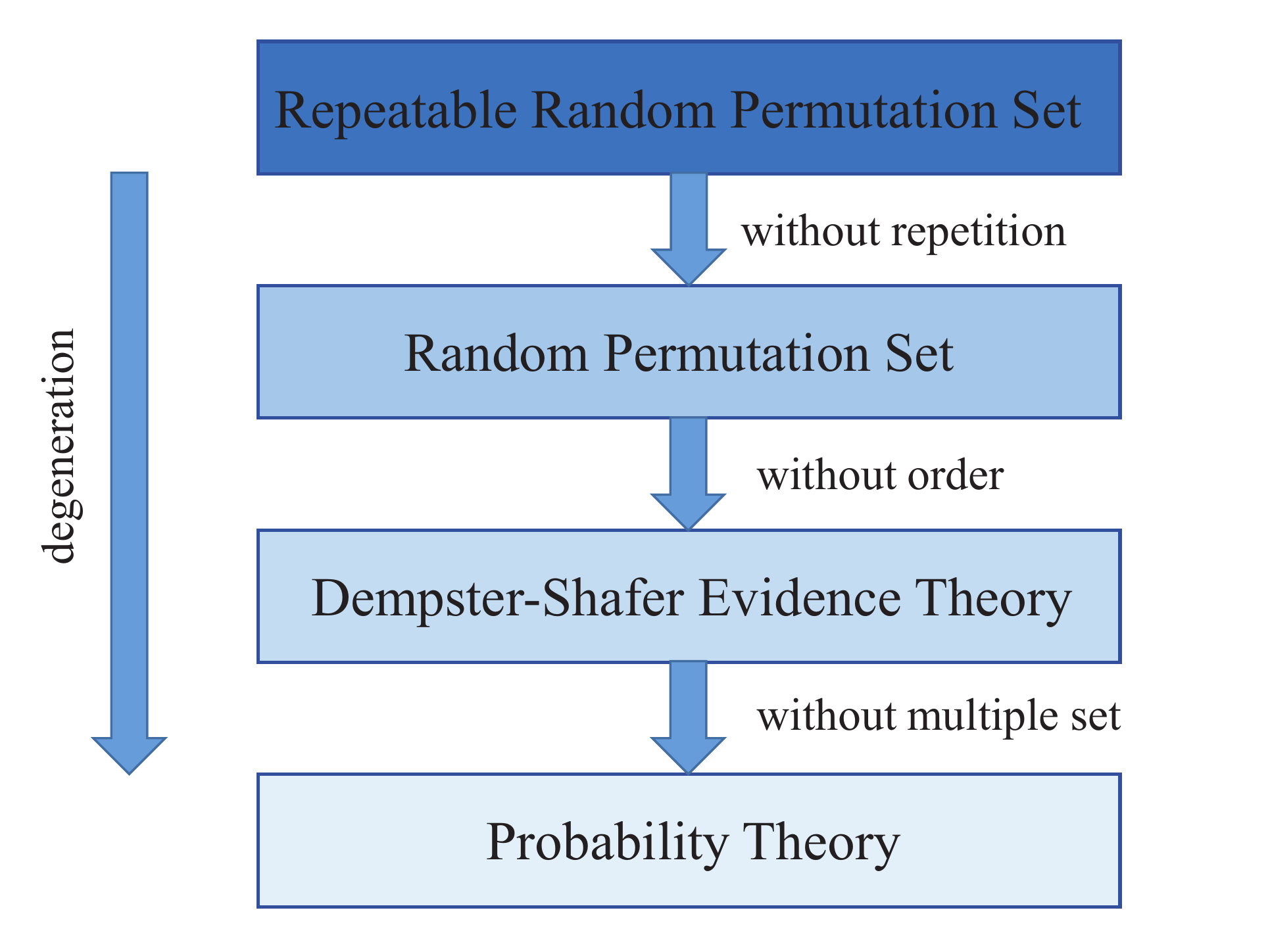}
	\caption{The degeneration relationships between probability theory, evidence theory, random permutation set and repeatable random permutation set}
	\label{fig:pdrr_2}
\end{figure}

In this paper, a new variation of RPS, Repeatable Random Permutation Set ($\rm R^2PS$) is developed. As shown in Figure \ref{fig:pdrr_2}, $\rm R^2PS$ could degenerate to RPS like RPS degenerates to DST. The event space, mass function and combination rules are defined based on RPS. Different from RPS, repetition of elements is allowed in these definitions. Besides, properties of left and right junctional sum combination rules including consistency, pseudo-Matthew effect and associativity are discussed. Based on these properties, an example of application in decision support system is demonstrated. 

The rest of this paper is organized as follows. In the next section, some basic definitions in RPS are shortly summarized. In Section 3, event space, mass function and combination rules are defined respectively. Furthermore, some properties of combination rules are discussed in Section 4. In section 5, an application example of supporting decision making is given.

\section{Preliminaries}
\label{Rreparatorywork}
In this section, we briefly introduced the event space, mass function and combination rules of RPS \cite{deng2022random}.

\begin{definition}[Permutation event space]
\label{PES}
The permutation event space (PES) consists of all permutations of the frame of discernment $\Omega=\{\omega_1,\omega_2,\cdots,\omega_N\}$. 
\begin{equation}
    \begin{aligned}
    PES(\Omega) = &\{ \phi,(\omega_1),(\omega_2),\cdots,(\omega_N),(\omega_1,\omega_2),(\omega_2,\omega_1 ),\cdots,(\omega_{N-1},\omega_N), \\ &(\omega_N,\omega_{N-1}),\cdots,
(\omega_1,\omega_2,\cdots,\omega_N),\cdots,(\omega_N,\omega_{N-1},\cdots,\omega_1) \}
    \end{aligned}
\end{equation}
\end{definition}

\begin{definition}[Permutation mass function]
\label{PMF}
Given a frame of discernment $\Omega$, a permutation mass function (PMF) is a mapping from PES to [0,1], denoted as $\mathscr{M}$.  
\begin{equation}
    \mathscr{M}: PES(\Omega) \rightarrow [0,1]
\end{equation}
\end{definition}

\begin{definition}[Left and right orthogonal sum]
\label{LRorthogonals}
Given two $\mathscr{M}$, the left and right orthogonal sum produce different combination results. The left orthogonal sum $\mathscr{M}^{L}$ preserves the order of the left $\mathscr{M}$ in case of conflict. Similarly, the right orthogonal sum $\mathscr{M}^{R}$ holds for the right.
\begin{equation}
\label{lorthosum}
    \mathscr{M}^{L}(A)=\left\{\begin{array}{l}
    \frac{1}{1-\overleftarrow{K}} \sum_{B \overleftarrow{\Pi} C=A} \mathscr{M}_{1}(B) \mathscr{M}_{2}(C), A \neq \varnothing \\ 
    0, A=\varnothing
    \end{array}\right.
\end{equation}
\begin{equation}
\label{rorthosum}
    \mathscr{M}^{R}(A)=\left\{\begin{array}{l}
    \frac{1}{1-\overrightarrow{K}} \sum_{B \overrightarrow{\Pi} C=A} \mathscr{M}_{1}(B) \mathscr{M}_{2}(C), A \neq \varnothing \\
    0, A=\varnothing
    \end{array}\right.
\end{equation}
where $\overleftarrow{\Pi}$ denotes left intersection and $\overrightarrow{\Pi}$ denotes right intersection. 
\begin{equation}
    M\overleftarrow{\Pi} N = M \verb|\\| \bigcup_{\omega \in M, \omega \notin N} \{\omega\}
\end{equation}
\begin{equation}
    M\overrightarrow{\Pi} N = N \verb|\\| \bigcup_{\omega \in N, \omega \notin M} \{\omega\}
\end{equation}
$\overleftarrow{K}$ and $\overrightarrow{K}$ denote the conflict coefficient.
\begin{equation}
    \overleftarrow{K} = \sum_{B \overleftarrow{\Pi} C=\varnothing} \mathscr{M}_1(B)\mathscr{M}_2(C)
\end{equation}
\begin{equation}
    \overrightarrow{K} = \sum_{B \overrightarrow{\Pi} C=\varnothing} \mathscr{M}_1(B)\mathscr{M}_2(C)
\end{equation}
\end{definition}

\section{Repeatable Random Permutation Set}
In this section, some basic concepts including event space, mass function and combination rules are defined.
\begin{definition}[Repeatable permutation event space]
Given a set $\Omega$, $|\Omega|=M$, $\Omega=\{ \omega_1, \omega_2, \cdots, \omega_M \}$, the corresponding repeatable permutation event space ($\rm R^2ES$) is 
\begin{equation}
\label{R2ES}
\begin{aligned}
R^2ES (\Omega) =& \{ O_{ij}|\ i = 1, 2,\cdots,N,j= 1,2,\cdots,M^i\}\\
 = &\{(\omega_1),(\omega_2),\cdots,(\omega_M),(\omega_1,\omega_1),(\omega_1,\omega_2),\cdots,(\omega_M,\omega_{M-1}),\\  & (\omega_M,\omega_M),\cdots (\omega_1,\omega_1,\cdots,\omega_1),\cdots,(\omega_N,\omega_N,\cdots,\omega_N)\},
\end{aligned}
\end{equation}
where $O_{ij}$ denotes the permutation event in $\rm R^2ES$. The permutation event is actually a tuple with elements from $\Omega$. The index $i$ indicates the cardinal and index $j$ differentiates events with the same cardinal $i$. The index $j$ ranges from $1$ to $M^i$.
\end{definition}

\begin{figure}
    \centering
    \includegraphics[width=12cm,height=7cm]{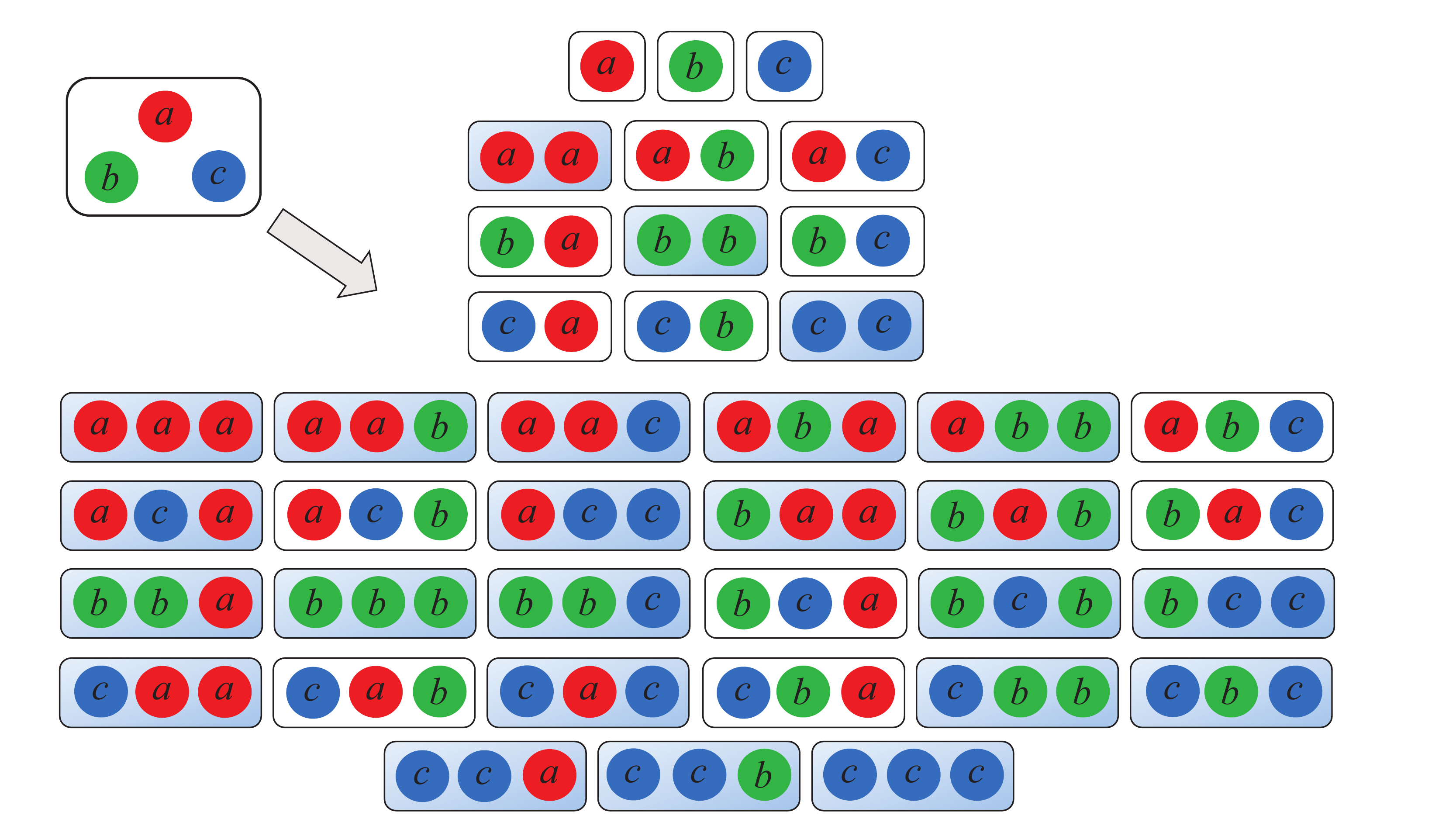}
    \caption{The demonstration of all possible repeatable permutation events in Example \ref{eg1:R2ES}}
    \label{fig:R2ES}
\end{figure}

\begin{figure}
	\centering
	\includegraphics[height=4.3cm]{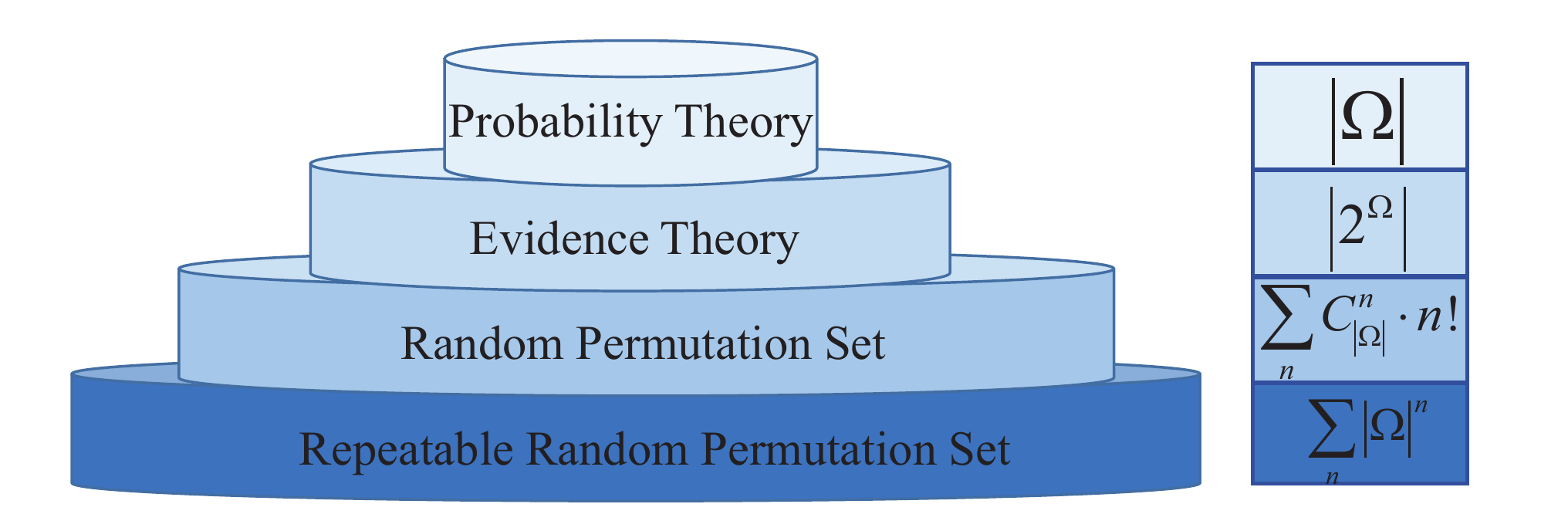}
	\caption{The enlargement of event space through probability theory, evidence theory, random permutation set and repeatable random permutation set}
	\label{fig:pdrr_1}
\end{figure}

Here we give an example to illustrate the $\rm R^2ES$.
\begin{example}
\label{eg1:R2ES}
Given a box with red, green and blue balls, the number of each color ball is large enough to be considered as infinite. The three colors are marked as $a$,$b$ and $c$ respectively. Hence, the frame of discernment is $\Omega=\{a,b,c\}$. Then, we draw balls from the box randomly without returning them. The number of balls we take from the box is no larger than 3. 

For instance, when we take two balls in sequence from the box, one possible scenario is we take a red ball firstly, and another red ball secondly, denoted as $(a,a)$, which is not considered in original RPS. Besides, as discussed in RPS, $(a,b)$ and $(b,a)$ are two different permutation events.

Then, all possible permutation events consist the $\rm R^2ES$ are shown in Equation \ref{eq:R2ES} and Figure \ref{fig:R2ES}, where the events in shadow are not contained in RPS.
\begin{equation}
\label{eq:R2ES}
\begin{aligned}
R^2ES(\Omega)=&\{ (O_{1,1}),(O_{1,2}),(O_{1,3}),\\
             &(O_{2,1}),(O_{2,2}),\cdots,(O_{2,9}), \\
             &(O_{3,1}),(O_{3,2}),\cdots,(O_{3,27})\} \\
=& \{  (a),(b),(c),\\
 & (a,a), (a,b),\cdots, (c,c),\\ 
 & (a,a,a), (a,a,b),\cdots, (c,c,c)\}
\end{aligned}
\end{equation}
\end{example}
Example \ref{eg1:R2ES} illustrates that all possible permutations are included in $\rm R^2ES$. As shown in figure \ref{fig:pdrr_1}, $\rm R^2ES$ is larger than the event space PES of RPS because the repetition in permutation is added to $\rm R^2ES$. 

\begin{definition}[Repeatable permutation mass function]
\label{RPMF}
Given a frame of discernment $\Omega$, a repeatable permutation mass function ($\rm R^2MF$) is a mapping from $\rm R^2ES$ to [0,1], denoted as  $\mathfrak{M}$.  
\begin{equation}
\mathfrak{M}: R^2ES(\Omega) \rightarrow [0,1]
\end{equation}
\end{definition}
Now, let us continue the ball-drawing model in Example \ref{eg1:R2ES}.
\begin{example}
\label{eg2:R2MF}
In the ball-drawing model with $\Omega=\{a,b,c\}$, one possible $\rm R^2MF$ is
\begin{equation}
\label{eq:eg2:R2MF}
    \mathfrak{M}(\Omega)=\{\left<(a,b),0.2\right>,\left<(b,b),0.3\right>,\left<(c,a),0.3\right>,\left<(a,a,c),0.2\right>\},
\end{equation}
in which $\left<(b,b),0.3\right>$ indicates that the $\rm R^2MF$ of event $(b,b)$ is 0.3, that is, the mass of both the first and second randomly chosen balls are green is 0.3. 
$\left<(c,a),0.3\right>$ represents that the $\rm R^2MF$ of event $(c,a)$ is 0.3. $(c,a)$ means one blue ball is taken firstly and a red ball is taken secondly. Compared with $\left<(a,a,c),0.2\right>$, although both events take red and blue balls, the order and frequency are different. It demonstrates that for RRPS, the difference in order and frequency is involved to distinguish events.
The mass associated with an event shows the chance of the event happening. From Equation \ref{eq:eg2:R2MF}, we can get the information that when the second ball is green, the chance of drawing the blue ball firstly is larger than red.
\end{example}
Example \ref{eg2:R2MF} demonstrates that the repeat of events is allowed, such as $\left<(b,b),0.3\right>$ and $\left<(a,a,c),0.2\right>$. Besides, the mass along with events indicates chance of happening.

\begin{definition}[Junctions]
Given two events $M$, $N$ in $\rm R^2ES$, the junctions between them are defined as left junction (LJ) and right junction (RJ).
\begin{equation}
    M\overleftarrow{\Pi} N = M \verb|\\| \bigcup_{\omega \in M, \omega \notin N} \{\omega\}\ \ \ \ (LJ)
\end{equation}
\begin{equation}
    M\overrightarrow{\Pi} N = N \verb|\\| \bigcup_{\omega \in N, \omega \notin M} \{\omega\}\ \ \ \ (RJ)
\end{equation}
where $\verb|\\|$ denotes remove elements.
\end{definition}
In descriptive language, LJ removes elements not in the right event $N$ from the left event $M$ and RJ removes element not in the left event $M$ from the right event $N$. For the better understanding, we give the following example.
\begin{example}
\label{eg:junction}
Considering two $\rm R^2MF$s $\mathfrak{M}_1$ and $\mathfrak{M}_2$, the corresponding R2MSs are
\begin{equation}
    \mathfrak{M}_1(\Omega) = \{\left<(a,b),0.4\right>,\left<(b),0.3\right>,\left<(a,a,c),0.3\right>\}
\end{equation}
\begin{equation}
    \mathfrak{M}_2(\Omega) = \{\left<(b,a),0.2\right>,\left<(b,b),0.3\right>,\left<(c,a),0.3\right>,\left<(a,c,c),0.2\right>\}
\end{equation}

The left and right joints of RRPS are illustrated as follows. 
\begin{equation}
(a,b)\overleftarrow{\Pi} (b,a) = (a,b) \verb|\\| \phi = (a,b)    
\end{equation}
\begin{equation}
(a,b)\overrightarrow{\Pi} (b,a) = (b,a) \verb|\\| \phi = (b,a)  
\end{equation}
The LJ of $(a,b)$ and $(b,a)$ is $(a,b)$, where two events contain the same elements. Hence, there is no element to remove and LJ is the same to the left event. The same goes for RJ.
\begin{equation}
(a,a,c)\overleftarrow{\Pi} (b,a) = (a,a,c) \verb|\\| (c) = (a,a)
\end{equation}
\begin{equation}
(a,a,c)\overrightarrow{\Pi} (b,a) = (b,a) \verb|\\| (b) = (a)
\end{equation}
The LJ and RJ of $(a,a,c)$ and $(b,a)$ both only contain element $(a)$ because $(a)$ is the only element contained in both events. However, the frequencies of $(a)$ differ in LJ or RJ, coinciding with left or right events, respectively.

The above discusses some instances of LJ and RJ. Then, all possible junctions between $\mathfrak{M}_1$ and $\mathfrak{M}_2$ are listed in Table \ref{tab:eg3:LJRJ}.
\begin{table}[h]
    \centering
    \caption{All possible junctions in $\mathfrak{M}_1$ and $\mathfrak{M}_2$}
    \label{tab:eg3:LJRJ}
    \begin{tabular}{cccc|cccc|cccc}
\hline																							
$\mathfrak{M}_1$	&	$\mathfrak{M}_2$	&	LJ	&	RJ	&	$\mathfrak{M}_1$	&	$\mathfrak{M}_2$	&	LJ	&	RJ	&	$\mathfrak{M}_1$	&	$\mathfrak{M}_2$	&	LJ	&	RJ	\\
\hline																							
\multirow{4}{*}{$(a,b)$}	&	$(b,a)$	&	$(a,b)$	&	$(b,a)$	&	\multirow{4}{*}{$(b)$}	&	$(b,a)$	&	$(b)$	&	$(b)$	&	\multirow{4}{*}{$(a,a,c)$}	&	$(b,a)$	&	$(a,a)$	&	$(a)$	\\
	&	$(b,b)$	&	$(b)$	&	$(b,b)$	&		&	$(b,b)$	&	$(b)$	&	$(b,b)$	&		&	$(b,b)$	&	$\phi$	&	$\phi$	\\
	&	$(c,a)$	&	$(a)$	&	$(a)$	&		&	$(c,a)$	&	$\phi$	&	$\phi$	&		&	$(c,a)$	&	$(a,a,c)$	&	$(c,a)$	\\
	&	$(a,c,c)$	&	$(a)$	&	$(a)$	&		&	$(a,c,c)$	&	$\phi$	&	$\phi$	&		&	$(a,c,c)$	&	$(a,a,c)$	&	$(a,c,c)$	\\
\hline																		
    \end{tabular}
\end{table}
\end{example}

Example \ref{eg:junction} demonstrates the process and results of right and left conjunction. 

\begin{definition}[left and right junctional sum]
\label{def:juncsum}
Given two $\rm R^2MF$s $\mathfrak{M}_1$ and $\mathfrak{M}_2$, the left junctional sum is denoted as $\mathfrak{M}_1 \overleftarrow{\oplus} \mathfrak{M}_2 $ and shorten as $\mathfrak{M}_{12}$.
\begin{equation}
    \mathfrak{M}_{12} (A) = \left\{\begin{array}{l}
    \frac{1}{1-\overleftarrow{K}} \sum_{B \overleftarrow{\Pi} C=A} \mathfrak{M}_{1}(B) \mathfrak{M}_{2}(C), A \neq \varnothing \\ 
    0, A=\varnothing
    \end{array}\right.
\end{equation}
\begin{equation}
    \overleftarrow{K} = \sum_{B \overleftarrow{\Pi} C=\varnothing} \mathfrak{M}_1(B)\mathfrak{M}_2(C)
\end{equation}
The right junctional sum is denoted as $\mathfrak{M}_1 \overrightarrow{\oplus} \mathfrak{M}_2 $ and shorten as $\mathfrak{M}_{21}$.
\begin{equation}
    \mathfrak{M}_{21} (A) = \left\{\begin{array}{l}
    \frac{1}{1-\overrightarrow{K}} \sum_{B \overrightarrow{\Pi} C=A} \mathfrak{M}_{1}(B) \mathfrak{M}_{2}(C), A \neq \varnothing \\ 
    0, A=\varnothing
    \end{array}\right.
\end{equation}
\begin{equation}
    \overrightarrow{K} = \sum_{B \overrightarrow{\Pi} C=\varnothing} \mathfrak{M}_1(B)\mathfrak{M}_2(C)
\end{equation}
\end{definition}

Continuing the $\rm R^2MF$s in Example \ref{eg:junction}, we provide the junctional sum of them.
\begin{example}
\label{eg:juncsum}
Take the calculation process of $\mathfrak{M}_{12}(b)$  as an instance,
\begin{equation}
\begin{aligned}
    \overleftarrow{K} =& \sum_{B \overleftarrow{\Pi} C=\phi} \mathfrak{M}_{1}(B) \mathfrak{M}_{2}(C) \\
    =& \mathfrak{M}_{1}(b) \mathfrak{M}_{2}(c,a) + \mathfrak{M}_{1}(b) \mathfrak{M}_{2}(a,c,c) + \mathfrak{M}_{1}(a,a,c) \mathfrak{M}_{2}(b,b) \\
    =& 0.3\times 0.3 + 0.3\times 0.2 + 0.3\times 0.3 = 0.24
\end{aligned}
\end{equation}
\begin{equation}
\begin{aligned}
        \mathfrak{M}_{12}(b) = & \frac{1}{1-\overleftarrow{K}} \sum_{B \overleftarrow{\Pi} C=b} \mathfrak{M}_{1}(B) \mathfrak{M}_{2}(C) \\
        = & \frac{1}{1-\overleftarrow{K}} (\mathfrak{M}_{1}(a,b) \mathfrak{M}_{2}(b,b)+\mathfrak{M}_{1}(b) \mathfrak{M}_{2}(b,a)+\mathfrak{M}_{1}(b) \mathfrak{M}_{2}(b,b)) \\
        = & \frac{1}{1-0.24} (0.4\times0.3+0.3\times0.2+0.3\times0.3) = 0.355
\end{aligned}
\end{equation}

The left junctional sum of $\mathfrak{M}_1$ and $\mathfrak{M}_2$ in Example \ref{eg:junction} is
\begin{equation}
\begin{aligned}
    \mathfrak{M}_{12}(\Omega) = \{ \left<(a,b),0.105\right>, \left<(b),0.355\right>, \left<(a),0.263\right>,\\ \left<(a,a),0.079\right>,\left<(a,a,c),0.197\right> \}.
\end{aligned}
\end{equation}
The right junctional sum is 
\begin{equation}
\begin{aligned}
    \mathfrak{M}_{21}(\Omega) = \{ \left<(b,a),0.105\right>, \left<(b,b),0.276\right>,\left<(b),0.079\right>, \\ \left<(a),0.342\right>, \left<(c,a),0.118\right>,\left<(a,c,c),0.079\right> \}.
\end{aligned}
\end{equation}
\end{example}

\section{Properties and numeral examples}
\label{sec_property}
In this section, we provide some examples to show some properties of RRPS. 
\begin{property}[Consistency]
The left and right junctional sum of two $\rm R^2MF$s $\mathfrak{M}_{1}$ and $\mathfrak{M}_{2}$ are consistent as shown below:
\begin{equation}
    \sum_{A^\downarrow=C} \mathfrak{M}_{12}(A) = \sum_{B^\downarrow=C} \mathfrak{M}_{21}(B) = m_{1\oplus 2}(C),
\end{equation}
where $A^\downarrow$ denotes the set consisting of all elements in $A$, which discarded the order and frequency information in $A$. To demonstrate this property, here we give an example.
\end{property}
\begin{example}
\label{eg:p1}
\begin{table}[]
    \centering
    \caption{An example to demonstrate the consistency property of RRPS}
    \label{tab:consistency}
    \begin{tabular}{c|cc|cc| cc|ccc}
\hline																			
$\rm R^2MF$s	&	$a$	&	$aa$	&	$ab$	&	$ba$	&	$b$	&	$bb$	&	$ca$	&	$acc$	&	$aac$	\\
\hline																			
$\mathfrak{M}_1$	&	$\cdot$	&	$\cdot$	&	0.4	&	$\cdot$	&	0.3	&	$\cdot$	&	$\cdot$	&	$\cdot$	&	0.3	\\
$\mathfrak{M}_2$	&	$\cdot$	&	$\cdot$	&	$\cdot$	&	0.2	&	$\cdot$	&	0.3	&	0.3	&	$\cdot$	&	0.2	\\
$m_1$	&	\multicolumn{2}{c|}{$\cdot$}			&	\multicolumn{2}{c|}{0.4}			&	\multicolumn{2}{c|}{0.3}			&	\multicolumn{3}{c}{0.3}					\\
$m_2$	&	\multicolumn{2}{c|}{$\cdot$}			&	\multicolumn{2}{c|}{0.2}			&	\multicolumn{2}{c|}{0.3}			&	\multicolumn{3}{c}{0.5}					\\
$\mathfrak{M}_1 \overleftarrow{\oplus} \mathfrak{M}_2 $	&	0.263	&	0.079	&	0.105	&	$\cdot$	&	0.355	&	$\cdot$	&	$\cdot$	&	$\cdot$	&	0.197	\\
$\mathfrak{M}_1 \overrightarrow{\oplus} \mathfrak{M}_2 $	&	0.342	&	$\cdot$	&	$\cdot$	&	0.105	&	0.079	&	0.276	&	0.118	&	$\cdot$	&	0.079	\\
$m_1 \oplus m_2$	&	\multicolumn{2}{c|}{0.342}			&	\multicolumn{2}{c|}{0.105}			&	\multicolumn{2}{c|}{0.355}			&	\multicolumn{3}{c}{0.197}					\\
\hline														
    \end{tabular}
\end{table}
Reviewing $\mathfrak{M}_{12}$ and $\mathfrak{M}_{21}$ in Example \ref{eg:juncsum}, the following relationships could be concluded. 
\begin{equation}
\begin{aligned}
        &\mathfrak{M}_{12}(a,b) = \mathfrak{M}_{21}(b,a) =0.105 = m_{1\oplus 2}(a,b)\\
    &\mathfrak{M}_{12}(b) = \mathfrak{M}_{21}(b) +\mathfrak{M}_{21}(b,b) =0.355 = m_{1\oplus 2}(b) \\
    &\mathfrak{M}_{12}(a)+\mathfrak{M}_{12}(a,a) = \mathfrak{M}_{21}(a) =0.342 = m_{1\oplus 2}(a) \\
    &\mathfrak{M}_{12}(a,a,c)  = \mathfrak{M}_{21}(c,a)+ \mathfrak{M}_{21}(a,c,c) = 0.197= m_{1\oplus 2}(a,c)
    \nonumber
\end{aligned}
\end{equation}
\end{example}
Example \ref{eg:p1} illustrates that the difference between left and right junctional sum lies in order, frequency and mass. Besides, the sum of masses of the same elements regardless of order and frequency are the same in left and right junctional sum. The consistency of the left and right sums reflects the consistency of fusion, and the fusion results are consistent on the whole. The difference is caused by the selective retention of order and frequency information under the premise of consistency.


\begin{property}[Pseudo-Matthew effect]
Given a sequence of $\rm R^2MF$s, combine them with left (right) junction. The frequency and order of elements in combined $\rm R^2MF$ are only determined by the first (last) event. 
\end{property}
The one-determined feature is similar to the Matthew effect. Hence, this property is named as Pseudo-Matthew effect. 
Take a simple example, given three $\rm R^2MF$s $\mathfrak{M}_1$, $\mathfrak{M}_2$ and $\mathfrak{M}_3$, $\mathfrak{M}_1 \overleftarrow{\oplus} \mathfrak{M}_2 \overleftarrow{\oplus} \mathfrak{M}_3 = \mathfrak{M}_1 \overleftarrow{\oplus} \mathfrak{M}_3 \overleftarrow{\oplus} \mathfrak{M}_2 $, that is, the right junctional combination result of a sequence of $\rm R^2MF$s is merely relevant to the first $\rm R^2MF$ and the order of rest $\rm R^2MF$s has no influence on the result. The same goes for the left junctional combination. 

Here, we give an explanation of the Pseudo-Matthew effect. According to the consistent property, the combination results of $n$ $\rm R^2MF$s in all possible different orders could degenerate into the same mass function in DST. The difference of combination results lies in the order and frequency of elements, which is totally determined by the left $\rm R^2MF$ through left junctional combination. In summary, the $m$ corresponding to the fusion result is determined after n $\rm R^2MF$s are selected, regardless of their order. The different order of fusion only affects the frequency and order of elements in $\rm R^2MF$ corresponding to $m$. When the first (last) $\rm R^2MF$s is given, the frequency and order of the elements are determined.



\begin{property}[Associativity]
Given three $\rm R^2MF$s $\mathfrak{M}_1$, $\mathfrak{M}_2$ and $\mathfrak{M}_3$, $(\mathfrak{M}_1 {\oplus} \mathfrak{M}_2) {\oplus} \mathfrak{M}_3$ is equal to $\mathfrak{M}_1 {\oplus} (\mathfrak{M}_2 {\oplus} \mathfrak{M}_3) $,
where ${\oplus}$ denote $\overleftarrow{\oplus}$ or $\overrightarrow{\oplus}$, \ie, all $\rm R^2MF$s are combined with one direction junctional sum.
\end{property}
Let ${\oplus}$ be $\overleftarrow{\oplus}$, the corresponding $m$ of $(\mathfrak{M}_1 \overleftarrow{\oplus} \mathfrak{M}_2) \overleftarrow{\oplus} \mathfrak{M}_3 $ and $\mathfrak{M}_1 \overleftarrow{\oplus} (\mathfrak{M}_2 \overleftarrow{\oplus} \mathfrak{M}_3) $ are same according to the consistent property. Besides, the order and frequency of elements in $(\mathfrak{M}_1 \overleftarrow{\oplus} \mathfrak{M}_2) \overleftarrow{\oplus} \mathfrak{M}_3 $ are determined by $\mathfrak{M}_1$. For $\mathfrak{M}_1 \overleftarrow{\oplus} (\mathfrak{M}_2 \overleftarrow{\oplus} \mathfrak{M}_3) $, the later could be seen as a whole and then $\mathfrak{M}_1 \overleftarrow{\oplus} (\mathfrak{M}_2 \overleftarrow{\oplus} \mathfrak{M}_3)$ is equivalent to $\mathfrak{M}_1 \overleftarrow{\oplus} \mathfrak{M}_4$, which indicates that
the frequency and order of elements in combination results are determined by $\mathfrak{M}_1$. Hence, not only the corresponding $m$, but also the the frequency and order of elements are the same. In conclusion, $(\mathfrak{M}_1 \overleftarrow{\oplus} \mathfrak{M}_2) \overleftarrow{\oplus} \mathfrak{M}_3 $ $=$ $\mathfrak{M}_1 \overleftarrow{\oplus} (\mathfrak{M}_2 \overleftarrow{\oplus} \mathfrak{M}_3) $. The same goes for the right junctional sum.

\begin{remark}
The left junctional sum of a sequence of RRPSs $\mathfrak{M}_1$, $\mathfrak{M}_2$, $\cdots$, $\mathfrak{M}_n$, $\mathfrak{M}_1 \overleftarrow{\oplus} \mathfrak{M}_2 \overleftarrow{\oplus} \cdots \mathfrak{M}_n = \mathfrak{M}_1 \overleftarrow{\oplus} \mathfrak{M}_k$. 

With the associativity property, the above equation holds if 
$\mathfrak{M}_k =\mathfrak{M}_1 \overleftarrow{\oplus} \mathfrak{M}_2 \overleftarrow{\oplus} \cdots \mathfrak{M}_n$, while this is a sufficient but not necessary condition.
With the consistency property, the above equation holds if $\sum_{A^\downarrow=C}\mathfrak{M}_k(A) = \sum_{A^\downarrow=C}\mathfrak{M}_1 \overleftarrow{\oplus}\mathfrak{M}_2 \overleftarrow{\oplus} \cdots \mathfrak{M}_n(A)$, \ie, $\mathfrak{M}_k$ is not necessarily equals to the left junctional sum of the last $(k-1)$ $\rm R^2MF$s. 
\end{remark}

\begin{remark}
The left junctional sum of a sequence of RRPSs $\mathfrak{M}_1$, $\mathfrak{M}_2$, $\cdots$, $\mathfrak{M}_n$, $\mathfrak{M}_1 \overleftarrow{\oplus} \mathfrak{M}_2 \overleftarrow{\oplus} \cdots \mathfrak{M}_n = \mathfrak{M}_1 \overleftarrow{\oplus} m_k$.

As a special case of $\mathfrak{M}_k$, $m_k(C)=\sum_{A^\downarrow=C}\mathfrak{M}_k(A)=\sum_{A^\downarrow=C}\mathfrak{M}_1 \overleftarrow{\oplus}\mathfrak{M}_2 \overleftarrow{\oplus} \cdots \mathfrak{M}_n(A)$. It verifies that the frequency and order of the elements are determined once the first (last) $\rm R^2MF$s is given. 
\end{remark}

\begin{remark}
The left junctional sum of a sequence of RRPSs $\mathfrak{M}_1$, $\mathfrak{M}_2$, $\cdots$, $\mathfrak{M}_n$ in different orders could generate at most $n$ different results.
\end{remark}

\begin{property}[Non-Associativity]
Given three $\rm R^2MF$s $\mathfrak{M}_1$, $\mathfrak{M}_2$ and $\mathfrak{M}_3$, $(\mathfrak{M}_1 {\oplus}_1 \mathfrak{M}_2) {\oplus}_2 \mathfrak{M}_3$ is not necessarily equal to $\mathfrak{M}_1 {\oplus}_1 (\mathfrak{M}_2 {\oplus}_2 \mathfrak{M}_3) $,
where ${\oplus}_1$ and ${\oplus}_2$ denote $\overleftarrow{\oplus}$ or $\overrightarrow{\oplus}$. 
\end{property}
Here an example is proposed to verify the non-associativity. 

\begin{example}
  Given three $\rm R^2MF$s $\mathfrak{M}_1$, $\mathfrak{M}_2$ and $\mathfrak{M}_3$, the combination results are shown in Table \ref{tab:eg:non-asso}. As listed in Table \ref{tab:eg:non-asso}, $\mathbf{(\mathfrak{M}_1 \overleftarrow{\oplus} \mathfrak{M}_2) \overrightarrow{\oplus} \mathfrak{M}_3}  $ is totally different with $\mathbf{\mathfrak{M}_1 \overleftarrow{\oplus} (\mathfrak{M}_2 \overrightarrow{\oplus} \mathfrak{M}_3)}  $.
\begin{table}[]
    \centering
    \caption{An example to demonstrate the non-associativity of RRPS}
    \label{tab:eg:non-asso}
    \begin{tabular}{ccccc ccccc}
\hline																			
$\rm R^2MF$s	&	$a$	&	$aa$	&	$ab$	&	$aac$	&	$acc$	&	$b$	&	$bb$	&	$ba$	&	$ca$	\\
\hline																			
$\mathfrak{M}_1$	&	$\cdot$	&	$\cdot$	&	0.4	&	0.3	&	$\cdot$	&	0.3	&	$\cdot$	&	$\cdot$	&	$\cdot$	\\
$\mathfrak{M}_2$	&	$\cdot$	&	$\cdot$	&	$\cdot$	&	$\cdot$	&	0.2	&	$\cdot$	&	0.3	&	0.2	&	0.3	\\
$\mathfrak{M}_3$	&	0.2	&	$\cdot$	&	$\cdot$	&	$\cdot$	&	0.3	&	$\cdot$	&	$\cdot$	&	0.5	&	$\cdot$	\\
$\mathfrak{M}_1 \overleftarrow{\oplus} \mathfrak{M}_2 $	&	0.263	&	0.079	&	0.105	&	0.197	&	$\cdot$	&	0.355	&	$\cdot$	&	$\cdot$	&	$\cdot$	\\
$\mathbf{(\mathfrak{M}_1 \overleftarrow{\oplus} \mathfrak{M}_2) \overrightarrow{\oplus} \mathfrak{M}_3}  $	&	\textbf{0.648}	&	$\cdot$	&	$\cdot$	&	$\cdot$	&	\textbf{0.072}	&	\textbf{0.216}	&	$\cdot$	&	\textbf{0.064}	&	$\cdot$	\\
$ \mathfrak{M}_2 \overrightarrow{\oplus} \mathfrak{M}_3  $	&	0.529	&	$\cdot$	&	$\cdot$	&	$\cdot$	&	0.177	&	0.177	&	$\cdot$	&	0.064	&	$\cdot$	\\
$\mathbf{\mathfrak{M}_1 \overleftarrow{\oplus} (\mathfrak{M}_2 \overrightarrow{\oplus} \mathfrak{M}_3)}  $	&	\textbf{0.384}	&	\textbf{0.264}	&	\textbf{0.064}	&	\textbf{0.072}	&	$\cdot$	&	\textbf{0.216}	&	$\cdot$	&	$\cdot$	&	$\cdot$	\\
\hline																			
    \end{tabular}
\end{table}
\end{example}

\section{Application in decision support system}
To verify the effectiveness of RRPS, a numerical application is demonstrated in this section. The following scenario is considered. A military agency needs to monitor certain airspace in real time. In the air space, there are three possible types of enemy aircraft including $a$,$b$ and $c$ and the frequency of each type of aircraft is unknown. In this agency, n experts estimate the type, order and frequency of enemy aircraft based on the data returned by the radar. 

Given sensor data, the first expert gives the assertion that $\mathfrak{M}_1(a,a)=0.6$, $\mathfrak{M}_1(a,b)=0.3$, $\mathfrak{M}_1(c)=0.1$. It indicates that the most possible condition is two type $a$ aircraft passing the airspace consequently. Then, the second possible condition is a type $b$ aircraft flying after a type $a$ aircraft. The possibility to be a type $c$ aircraft is small. 
Similarly, the latter two experts give assertions $\mathfrak{M}_2$ and $\mathfrak{M}_3$ shown in Table \ref{tab:app_expert}. 
\begin{table}[h]
    \centering
    \caption{The estimation produced by three experts}
    \label{tab:app_expert}
    \begin{tabular}{ccccc ccc}
\hline															
$\rm R^2MF$s	&	$a$	&	$aa$	&	$ab$	&	$b$	&	$ba$	&	$c$	&	$cba$	\\
\hline															
$\mathfrak{M}_1$	&	$\cdot$	&	0.6	&	0.3	&	$\cdot$	&	$\cdot$	&	0.1	&	$\cdot$	\\
$\mathfrak{M}_2$	&	0.4	&	$\cdot$	&	0.3	&	0.2	&	$\cdot$	&	$\cdot$	&	0.1	\\
$\mathfrak{M}_3$	&	$\cdot$	&	$\cdot$	&	0.6	&	$\cdot$	&	0.2	&	0.2	&	$\cdot$	\\
\hline															
\end{tabular}
\end{table}

Based on the consistent and Pseudo-Matthew effect property illustrated in Section \ref{sec_property}, there merely exist $n$ possible combination results of $n$ experts' assertions. All combination results are shown in Table \ref{tab:fusion}.
\begin{table}[h]
    \centering
    \caption{All possible fusion results of experts' assertions}
    \label{tab:fusion}
    \begin{tabular}{c|cc|c|c c|c}
\hline													
$\rm R^2MF$s	&	$a$	&	$aa$	&	$b$	&	$ab$	&	$ba$	&	$c$	\\
\hline													
$m_{23}$	&	\multicolumn{2}{c|}{0.43}			&	0.22	&	\multicolumn{2}{c|}{0.32}			&	0.03	\\
$m_{13}$	&	\multicolumn{2}{c|}{0.65}			&	$\cdot$	&	\multicolumn{2}{c|}{0.32}			&	0.03	\\
$m_{12}$	&	\multicolumn{2}{c|}{0.77}			&	0.09	&	\multicolumn{2}{c|}{0.13}			&	0.01	\\
$m_{123}$	&	\multicolumn{2}{c|}{0.78}			&	0.09	&	\multicolumn{2}{c|}{0.13}			&	0.01	\\
$\mathfrak{M}_1\overleftarrow{\oplus} m_{23}$	&	$\cdot$	&	0.78	&	0.09	&	0.13	&	$\cdot$	&	0.01	\\
$\mathfrak{M}_2\overleftarrow{\oplus} m_{13}$	&	0.78	&	$\cdot$	&	0.09	&	0.13	&	$\cdot$	&	0.01	\\
$\mathfrak{M}_3\overleftarrow{\oplus} m_{12}$	&	0.78	&	$\cdot$	&	0.09	&	0.10 	&	0.03	&	0.01	\\
\hline													
\end{tabular}
\end{table}
All possible combination results degenerate into the same $m$, \ie, the difference between results merely lies in the frequency and order of elements. The corresponding $m$ is calculated in Equation \ref{eq:am}.
\begin{equation}
\label{eq:am}
    m(A) = \sum_{B^\downarrow=A} \mathfrak{M}(B),
\end{equation}
Specifically, the $m$ for $\mathfrak{M}_1$, $\mathfrak{M}_2$ and $\mathfrak{M}_3$ is shown in Table \ref{tab:fusion} as $m_{123}$. 
Because of the consistent property, the decision process is to make decisions between events with the same elements in different frequencies and orders. Winner-take-all strategy is taken for decision-making. The most supported RRPS event obtains all mass of the corresponding element in $m$. As defined in Equation \ref{eq:a}, the most supported event $E$ is the event with the highest comprehensive support under different expert-oriented decisions. 
\begin{equation}
\label{eq:a}
    E = argmax \sum_{i = 1}^n \mathfrak{M}_{i\oplus}(C)\ \ for\ all\ C^\downarrow=A
\end{equation}
Hence, the decision based on experts' assertions is 
\begin{equation}
\mathfrak{M}_d(\Omega) = \{ \left<(a),0.78\right>,\left<(b),0.09\right>,\left<(ab),0.13\right>,\left<(c),0.01\right> \}.
\end{equation}
The most possible scenario is one type $a$ aircraft passing the airspace with the belief mass of 0.78. The secondly possible scenario is one type $b$ aircraft following one type $a$ aircraft passing the airspace with the possibility of 0.13. The probability of one type $b$, $c$ aircraft is 0.09, 0.01 respectively.

\section{Conclusions}
\label{conclusion}
In real-world, the repetition of events often exist in sequences of events, while it is not allow in RPS. To address this issue, we expand the notation of RPS into repeatable RPS which is able to suit more practical situations. The main contributions of this paper are as follows.
\begin{itemize}
\item After proposing the notation of Repeatable random permutation set (RRPS), the corresponding event space ($\rm R^2ES$) and mass function ($\rm R^2MF$) are defined. Besides, a ball-drawing model is demonstrated for straightforward understanding. 
\item The combination rules of $\rm R^2MF$ are defined as left and right junctional sum. Some properties of the combination rules including consistency, Pseudo-Matthew effect, Associativity and Non-Associativity are researched. 
\item The efficiency and effectiveness of RRPS are verified through a threat analysis application. Experts' opinions are combined by RRPS combination rules and decision is made based on all possible combination results.
\end{itemize}

\section*{Acknowledgments}
The work is partially supported by National Natural Science Foundation of China (Grant Nos. 61973332).
\section*{Conflict of interests}
The authors declare that there is no conflict of interests regarding the publication of this paper.


\bibliographystyle{elsarticle-num}
\bibliography{References}

\end{document}